%% file: main_arxiv.tex
\documentclass[10pt,twocolumn,letterpaper]{article}

\usepackage[pagenumbers]{iccv} 

\usepackage[accsupp]{axessibility}  


\definecolor{iccvblue}{rgb}{0.21,0.49,0.74}
\usepackage[pagebackref,breaklinks,colorlinks,allcolors=blue]{hyperref}

\usepackage{graphicx}
\usepackage{amsmath}
\usepackage{amssymb}
\usepackage{amsfonts}
\usepackage{booktabs}
\usepackage{color}
\usepackage{multirow}
\usepackage{subcaption}
\usepackage{caption}

\usepackage{url}
\usepackage{graphicx}
\usepackage{amsmath}
\usepackage{amssymb}
\usepackage{booktabs}
\usepackage{paralist}

\usepackage{multirow}
\usepackage{multicol}
\usepackage{booktabs}
\usepackage{amsmath}
\usepackage{array}
\usepackage{lipsum}
\usepackage{tabularx}
\usepackage{enumitem}
\usepackage{amsmath} 
\usepackage{enumitem}

\def\paperID{7340} 
\def\confName{ICCV}
\def\confYear{2025}

\title{Are They the Same? Exploring Visual Correspondence Shortcomings of Multimodal LLMs}

\author{
Yikang Zhou$^{1}$\textsuperscript{*}~\quad
Tao Zhang$^{1}$\textsuperscript{*}~\quad
Shilin Xu$^{3}$\quad
Shihao Chen$^{1}$\quad
Qianyu Zhou$^{5}$\quad
Yunhai Tong$^{3}$\quad\\
Shunping Ji$^{1}$\textsuperscript{\textdagger}~\quad
Jiangning Zhang$^{4}$\quad
Lu Qi$^{1}$\quad
Xiangtai Li$^{2}$\textsuperscript{\textdaggerdbl}\\
$^1$Wuhan University\quad
$^2$Bytedance Seed\quad
$^3$Peking University\quad
$^4$ Zhejiang University\quad
$^5$ SJTU\\
{\tt\small \{zhouyik,zhang\_tao,jishunping\}@whu.edu.cn, xiangtai94@gmail.com}\\
{\small{\url{https://zhouyiks.github.io/projects/CoLVA/}}}
}

\begin{document}
\twocolumn[{%
\renewcommand\twocolumn[1][]{#1}%
\maketitle
\vspace{-8mm}
\centering
\includegraphics[width=0.90\textwidth]{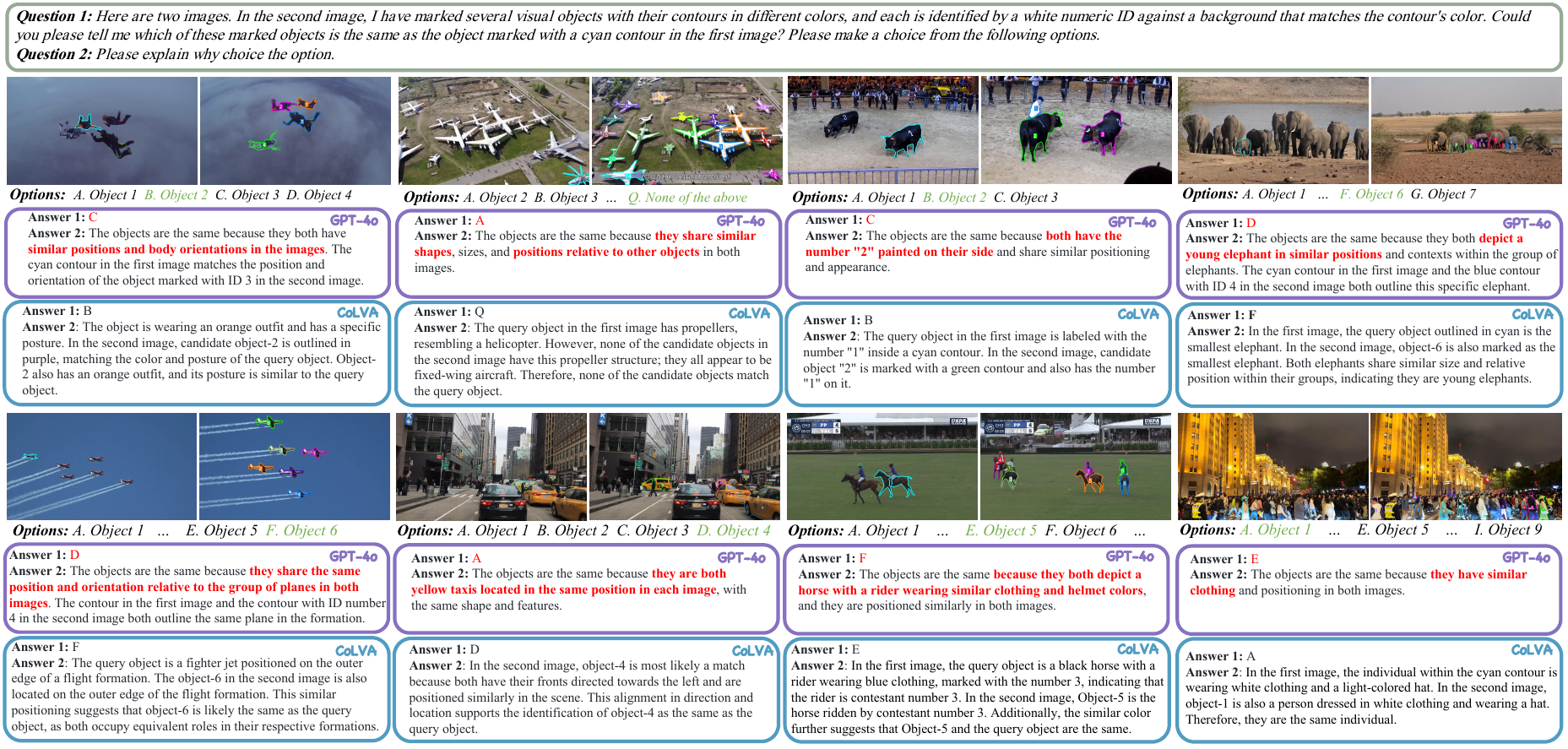}
\captionof{figure}{Visualization results of GPT-4o and our proposed CoLVA on challenging cases of MMVM benchmarks. The GPT-4o's answers are incorrect for all these examples, with the errors highlighted in red. The correct answers in the options are highlighted in green.}
\label{fig:teaser}
\vspace{0.7cm}
}]

\footnotetext{\textsuperscript{*}Equal contribution. \textsuperscript{\textdagger}Corresponding author. \textsuperscript{\textdaggerdbl}Project leader.}

\input{ICCV/latex/0_abs}

\input{ICCV/latex/1_intro}

\input{ICCV/latex/2_related_work}

\input{ICCV/latex/3_method}
\input{ICCV/latex/4_exp}

\input{ICCV/latex/5_conclusion}

\input{ICCV/latex/6_appendix}

\clearpage
\newpage
{
    \small
    \bibliographystyle{ieeenat_fullname}
    \bibliography{main}
}


\end{document}

%% file: ICCV/latex/0_abs.tex
\begin{abstract}
Recent advancements in multimodal large language models (MLLM) have shown a strong ability in visual perception, reasoning abilities, and vision-language understanding.
However, the visual matching ability of MLLMs is rarely studied, despite finding the visual correspondence of objects is essential in computer vision. 
Our research reveals that the matching capabilities in recent MLLMs still exhibit systematic shortcomings, even with current strong MLLMs models, GPT-4o. 
In particular, we construct a Multimodal Visual Matching (MMVM) benchmark to fairly benchmark over 30 different MLLMs. 
The MMVM benchmark is built from 15 open-source datasets and Internet videos with manual annotation.
%
%
%
In addition, we have designed an automatic annotation pipeline to generate the MMVM SFT dataset, including 220K visual matching data with reasoning annotation.
To our knowledge, this is the first visual corresponding dataset and benchmark for the MLLM community.
Finally, we present CoLVA, a novel contrastive MLLM with two novel technical designs: fine-grained vision expert with object-level contrastive learning and instruction augmentation strategy.
The former learns instance discriminative tokens, while the latter further improves instruction following ability.
CoLVA-InternVL2-4B achieves an overall accuracy (OA) of 49.80\% on the MMVM benchmark, surpassing GPT-4o and the best open-source MLLM, Qwen2VL-72B, by 7.15\% and 11.72\% OA, respectively. 
These results demonstrate the effectiveness of our MMVM SFT dataset and our novel technical designs.
Code, benchmark, dataset, and models will be released.
%
\end{abstract}

%% file: ICCV/latex/1_intro.tex
\section{Introduction}
\label{sec:intro}

MLLMs~\cite{Liu2023llava,liu2024llavanext,chen2024internvl,wang2024qwen2,openai2023gpt4} have made remarkable progress with the development of Large Language Models (LLMs)~\cite{llama,llama2, qwen2}. 
They have greatly benefited various applications, including image and video understanding~\cite{zhang2024omg, Liu2023llava, cai2024vipllava}, visual question answering (VQA)~\cite{chen2024internvl, wang2024qwen2}, and visual grounding~\cite{Lai2023lisa, rasheed2024glamm, zhang2024omg}.
Despite the advancements of MLLMs with various capabilities~\cite{cai2024vipllava,li2024llavanextinter,chen2024internvl, wang2024qwen2, li2024llamavid, song2024moviechat, hong20233d-llm}, they often struggle with visual correspondence, a fundamental ability that plays a key role in several vision tasks, including tracking~\cite{ravi2024sam2,li2024masa}, feature matching~\cite{baumberg2000reliable}, and reconstruction~\cite{geiger2011stereoscan}. 
As shown in Fig.~\ref{fig:teaser}, even the GPT-4o~\cite{openai2023gpt4} cannot understand some simple matching questions well. 
This limitation is critical, as it hinders MLLMs from comprehending correspondence-aware information.

Based on this motivation, we aim to systematically analyze this problem in MLLMs~\cite{Liu2023llava,openai2023gpt4,chen2024internvl,wang2024qwen2} and propose a corresponding method to address it. 
First, a new and challenging benchmark on instance-level correspondence across multiple images is required due to the lack of comprehensive evaluations for this direction. 
Specifically, we collect 1,510 samples from both 15 public video datasets~\cite{qi2022occluded, yang2019video, wang2024ov, ding2023mose, yu2020bdd100k, athar2023burst, cui2023sportsmot, Hu2023videocube, hou2020multiview, peng2024vasttrack, milan2016mot17, fan2019lasot, miao2022vipseg, chavdarova2018wildtrack, valmadre2018long} and internet video platforms. 
These samples encompass various scenes, including indoor environments, urban settings, cartoons, drone footage, and various social activity scenarios. 
Each sample is meticulously annotated with multi-image QA pairs by three skilled annotators. 
The diversity of these samples enables us to evaluate the visual matching capabilities of MLLMs across multiple dimensions of matching cues. 
We summarize eight types of matching cues (such as color, markers), which are the most frequently encountered by humans. (See the Sec.~\ref{sec:benchmark})

\begin{figure}[t]
\centering
\includegraphics[width=0.45\textwidth]{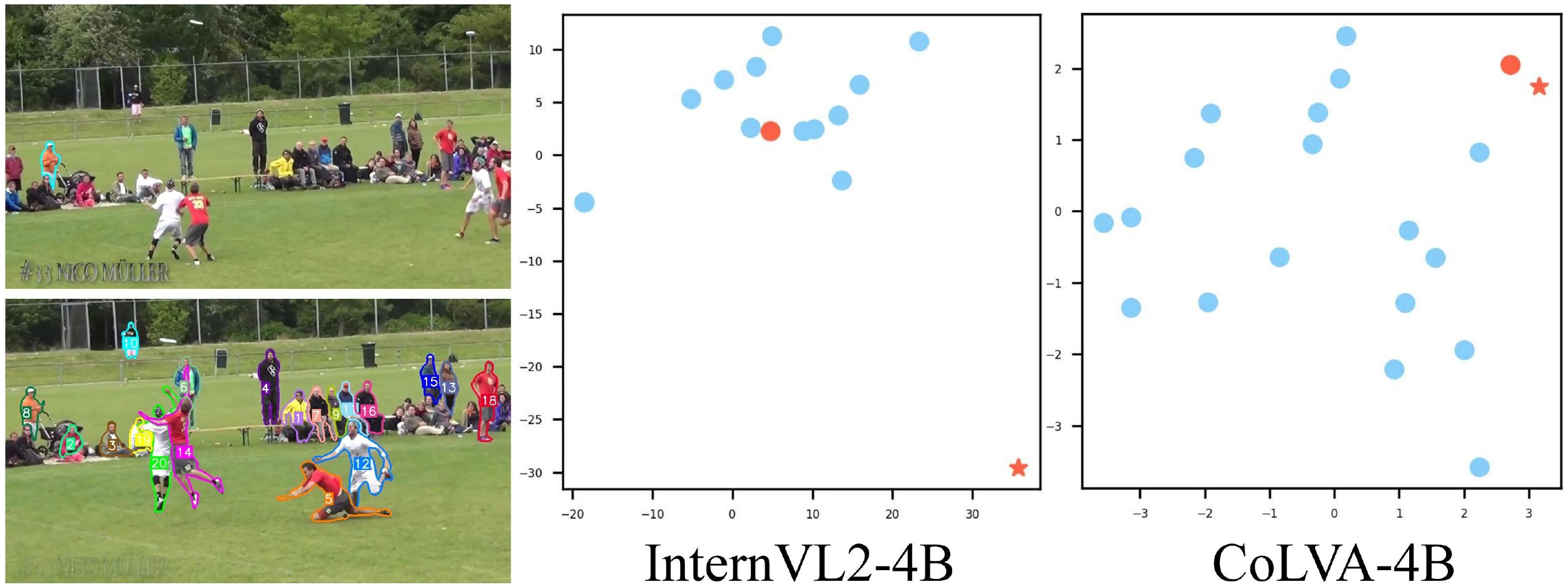}
\caption{\small The PCA visualization of learned object embeddings by InternVL2-4B and our CoLVA-4B. The object embeddings are obtained by applying average pooling to the visual tokens using mask annotations. The red star represents the query object in the first image. The red dot represents the matched target in the second image. The blues dots represent other candidates. More PCA visualizations can be found in the appendix.}
\label{fig:pca}
\vspace{-2mm}
\end{figure}

Then, we evaluate 36 state-of-the-art (SOTA) MLLM methods on our benchmark.
The quantitative evaluation in our benchmark highlights the merits of our work, as the strong model, Qwen2-VL-72B-Instruct~\cite{wang2024qwen2}, achieves only 38\% overall accuracy.
This indicates that current state-of-the-art (SOTA) MLLMs exhibit notable matching shortcomings. 
Through quantitative experiments and PCA visualization analysis (Fig.~\ref{fig:pca}), we identify two primary factors contributing to these visual shortcomings: \textit{1) Although current MLLMs possess a specific capability to recognize objects' appearances and positions, they lack the corresponding data to teach them how to utilize this foundational knowledge and these abilities for visual matching}; 
\textit{2) Current MLLMs rely on CLIP models to understand images and cannot comprehend fine-grained and discriminative visual features}.

%
These findings motivate us to develop an automatic data generation pipeline for building a high-quality visual matching SFT dataset (MMVM dataset). 
The MMVM dataset includes 220k multi-choice QA pairs. 
Each is accompanied by matching rationales.
We establish a simple yet effective baseline, CoLVA, and fine-tune it using our MMVM dataset. 
CoLVA integrates two simple yet effective techniques into existing SOTA MLLMs, such as InternVL2~\cite{chen2024internvl} and Qwen2VL~\cite{wang2024qwen2}, to enhance correspondence training: a fine-grained vision expert with object-level contrastive learning (OCL) and instruction augmentation (IA).
Specifically, we perform object-level contrastive learning between the MLLM visual encoder and the vision expert, motivated by previous works~\cite{fischer2023qdtrack,li2022video}.
This enables the vision expert to learn discriminative features within the semantic space of the MLLM. 
First, it preserves fine-grained visual features since our benchmark involves detailed visual prompts as inputs.
In addition, it achieves modality alignment through contrastive learning. 
This dual-purpose design highlights the novelty of our OCL strategy.
Furthermore, we integrate the learned discriminative object-level features into the instructions. 
This allows gradients to be directly backpropagated through the object-level features to the corresponding image features, enabling the MLLM to learn the required discriminative and fine-grained features more effectively. 
Moreover, this enhances our ability to refer to multiple objects within the images.
%
%
Finally, extensive experiments demonstrate the effectiveness of our MMVM dataset and network design. 
Our CoLVA-InternVL2-4B achieves improvements of 11.72\% and 7.15\% over the open-source Qwen2VL-72B and the proprietary GPT-4o, respectively.

%
To sum up, our contributions are four-fold:
\begin{compactitem}
    \item {We establish a challenging benchmark for the visual matching problem in Multimodal LLMs.}
    \item {We propose a high-quality MMVM dataset, which contains 220k matching QA pairs with reasoning texts.}
    \item {We propose two simple, yet effective techniques for correspondence learning.}
    \item {Extensive experiments demonstrate the effectiveness of our proposed dataset and technical contributions.}
\end{compactitem}

%% file: ICCV/latex/2_related_work.tex
\section{Related Work}
\label{sec:related_work}





\noindent
\textbf{Multi-modal Large Language Models.} With the development of LLMs~\cite{anil2023palm,Touvron2023llama,touvron2023llama2,team2023internlm,bai2023qwen,abdin2024phi}, Multimodal LLMs raise significant attention in image and video understanding. 
Current MLLMs~\cite{Liu2023llava,liu2024llava1.5,dai2023instructblip,chen2024internvl,wang2024qwen2} explore adapter layers to transfer visual features (CLIP~\cite{radford2021learning}) into visual token input for LLMs. 
LLaVA~\cite{Liu2023llava} is one representative work that uses MLPs as a visual adapter. 
The following works~\cite{liu2024llava1.5,li2024llavanextinter,chen2023sharegpt4v} mainly explore high-quality data for both pre-training and instruction tuning. 
Meanwhile, several works~\cite{cai2024vipllava,yuan2024osprey,wu2024controlmllm, lin2024mdvp} explore stronger visual cues or inject fine-grained visual prompts into MLLMs.
For example, VIP-LLaVA~\cite{cai2024vipllava} integrates arbitrary visual prompts into LLaVA~\cite{Liu2023llava}.
Several works have also studied MLLMs in video~\cite{zhang2023videollama,lin2023mmvid,zhao2023LaViLa,maaz2023video-chatgpt,song2024moviechat,lin2023videollava} and 3D~\cite{hong20233d-llm,xu2025pointllm,wang2023chat-3d}.
In particular, recent works on video MLLMs can be summarized in two directions. 
One direction~\cite{lin2023videollava,li2024llamavid} aims to compress visual tokens for longer video modeling. The other direction designs stronger memory attention to achieve state-of-the-art performance. 
Several works~\cite{wang2024groundedvideollm,qu2024chatvtg,sun2024chattracker} explore video grounding and provide strong text features for visual tracking. 
To our knowledge, no works explore fine-grained visual correspondence understanding in MLLMs.
Our work is the first step in a fine-grained correspondence understanding of multi-images.

\begin{figure}[t]
\centering
\includegraphics[width=0.45\textwidth]{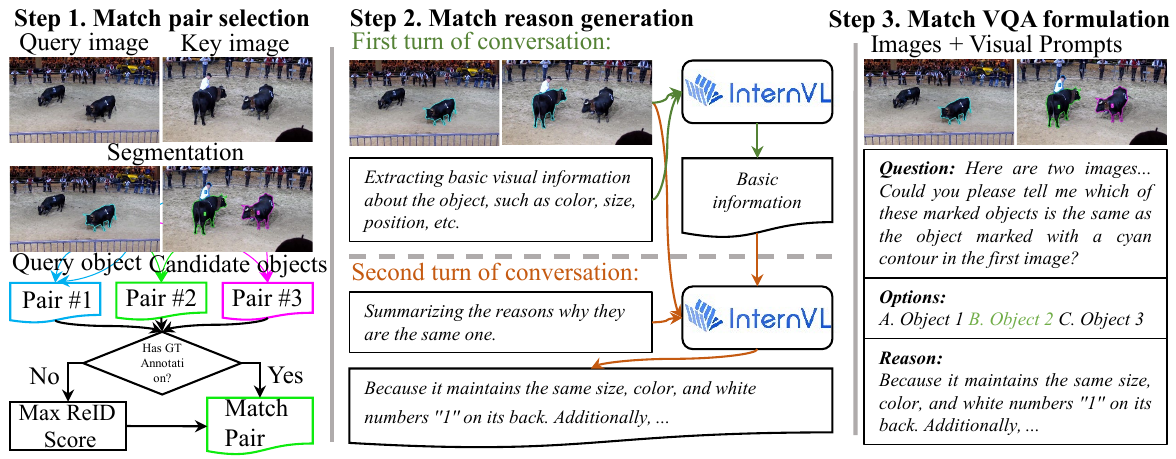}
\caption{\small The proposed automatic visual matching data generation pipeline. We begin by collecting various image pairs from open-source video datasets. We then utilize the InternVL-76B model to generate the reasons for object matching. Finally, we organize all the image pairs and the generated matching reasons into a unified format for multi-image VQA tasks.}
\label{fig:sft_data}
\end{figure}

\begin{figure}[t]
\centering
\includegraphics[width=0.45\textwidth]{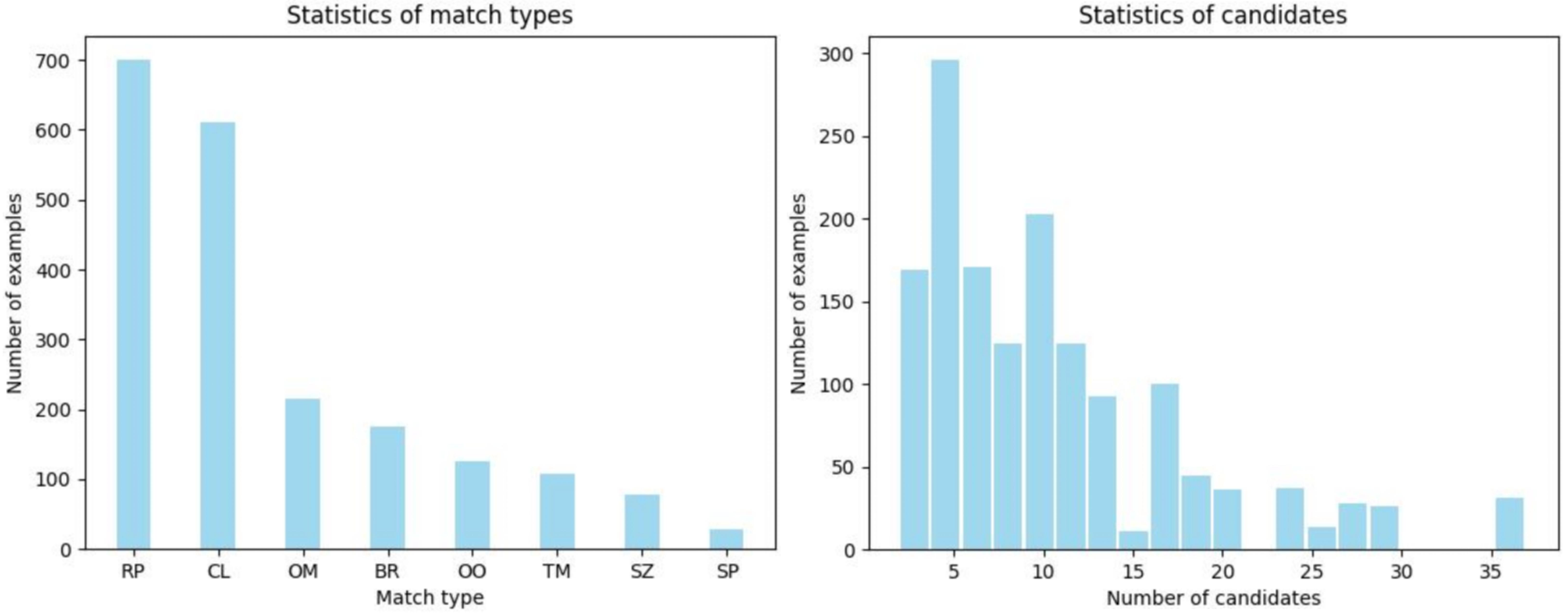}
\caption{\small The statistics of the MMVM benchmark. The left side presents the statistics of the example counts for matching cues, while the right side displays the statistics regarding the number of candidates in the examples.}
\label{fig:mmvm_statistics}
\end{figure}

\begin{figure*}[th]
\centering
\includegraphics[width=0.99\textwidth]{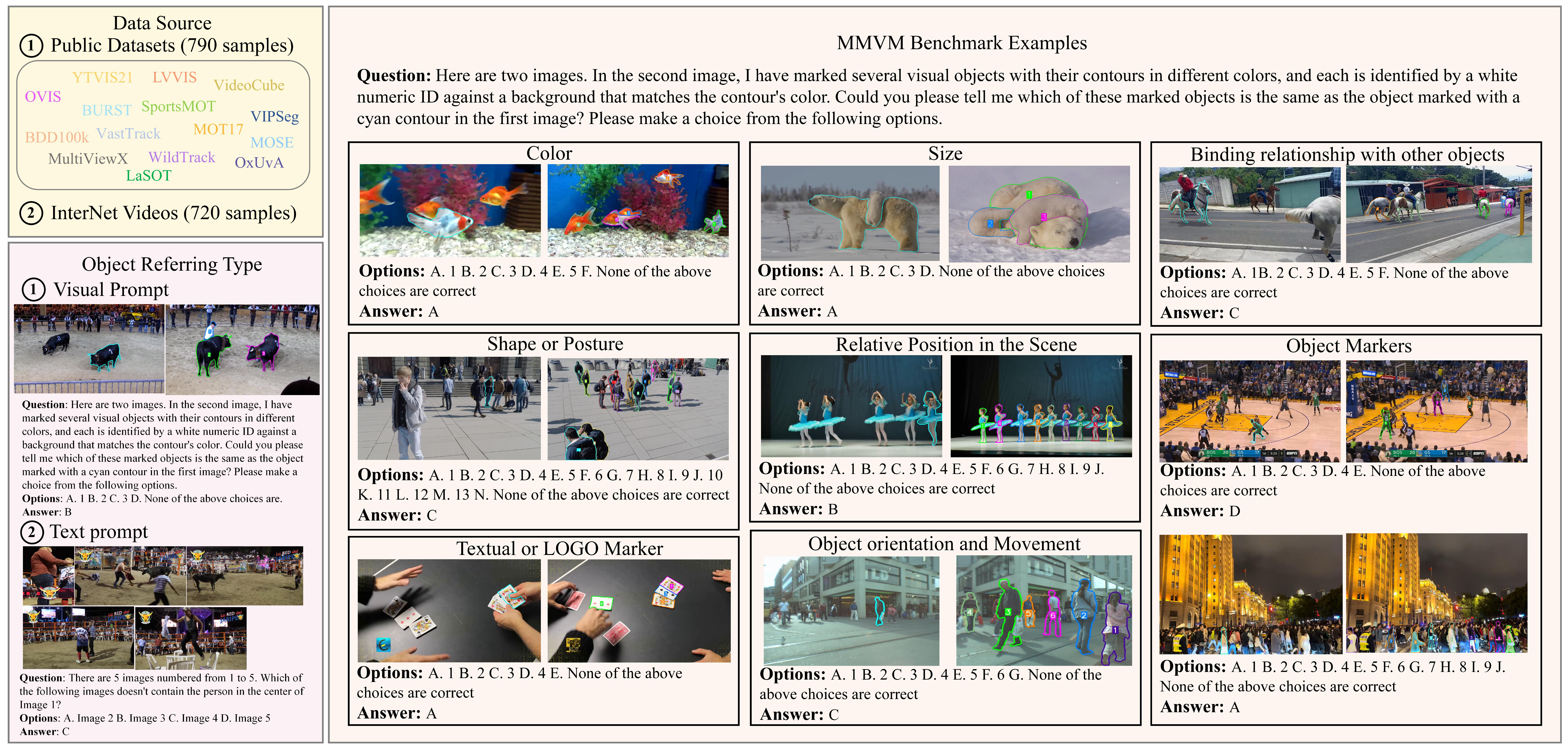}
\caption{\small Visualization of MMVM Benchmark. Our MMVM Benchmark contains 1,510 manually annotated multi-image QA pairs, 8 matching patterns, and 2 types of object referring methods. We collect the evaluation samples from 15 open-source video datasets and various internet video platforms.}\vspace{-3mm}
\label{fig:benchmark}
\end{figure*}

\noindent
\textbf{Visual Corresponding Learning.} Learning instance discriminative features is critical to many applications, including object tracking, person re-identification, and multi-view reconstruction. 
Several works~\cite{zolkepli2024mmmmodal,meng2024mmiu,liu2024llavanext,wang2024qwen2} explore the cross-image understanding of MLLMs, and most works follow the VQA pipeline. 
Our works are inspired by previous visual corresponding learning~\cite{wu2022defense, zhang2023dvis, zhang2023dvis++, li2024masa, li2023tube, zhou2024dvis, zhang20231stlavos, zhang20231stpvuw} and present a new learning framework with contrastive visual tokens for current MLLMs.

\noindent
\textbf{Region Understanding of Multimodal LLMs.} Understanding fine-grained information is also important to build stronger MLLMs. Several works~\cite{lin2024mdvp,yuan2024osprey,rasheed2024glamm,zhang2024omg} explore region-aware or mask-aware instruction tuning pipelines to MLLMs.
In particular, Osprey~\cite{yuan2024osprey} adopts mask-aware pooling into MLLMs to understand fine-grained region features.
Meanwhile, several works~\cite{rasheed2024glamm,zhang2024omg,Lai2023lisa} explore the visual grounding of MLLMs to make MLLMs output specific locations. 
GLaMM~\cite{rasheed2024glamm} combines interactive segmentation with LLaVA~\cite{Liu2023llava} and proposes grounded VQA and segmentation in one framework. 
Our studies explore region-level understanding in cross-image settings, which is orthogonal to previous works.

\noindent
\textbf{Evaluating Multimodal LLMs.} Earlier works mainly focus on traditional VQA queries in general cases, such as TextVQA~\cite{singh2019towards}, VQAv2~\cite{balanced_vqa_v2}, and GQA~\cite{Hudson2019gqa}. 
Recent works like MME~\cite{fu2024mme}, MM-VeT~\cite{yu2023mm-vet}, and MM-Bench~\cite{liu2025mmbench}, are designed to evaluate the specific features of MLLMs, including hallucination, reasoning, robustness, OCR, and chat analysis,
Meanwhile, several works~\cite{tong2024cambrian,tong2024eyes} explore the vision-centric features of MLLMs.
We argue that our benchmark is a solid complement to existing MLLMs, making current MLLMs understand fine-grained matching ability without degradation of VQA tasks.

%% file: ICCV/latex/3_method.tex
    
   
\section{MMVM Dataset and Benchmark}
\label{sec:benchmark}

We first introduce the strategy for constructing the MMVM dataset (detailed in Sec.~\ref{sec:train_data}). We then detail the MMVM benchmark in Sec.~\ref{sec:evaluate_data}.

\subsection{MMVM Dataset}
\label{sec:train_data}

To construct a large-scale visual matching dataset, we leverage existing video datasets to generate multiple-choice QA pairs (\textbf{Step 1}) and collect reasoning texts for visual matching by prompting advanced MLLMs (\textbf{Step 2}). Finally, we organize the multi-choice QA pairs and reasoning texts into a multi-turn dialogue format (\textbf{Step 3}).

\noindent\textbf{Multiple-choice QA Generation.} We filter and reorganize the train sets of current video segmentation datasets, including OVIS~\cite{qi2022occluded}, YouTube-VIS 2021~\cite{yang2019video}, VIPSeg~\cite{miao2022vipseg}, BDD100K~\cite{yu2020bdd100k}, and BURST~\cite{athar2023burst}. 
As illustrated on the left side of Fig.~\ref{fig:sft_data}, we sample frames at fixed 1-second intervals for each video and subsequently organize adjacent frames into image pairs. 
Each image pair contains multiple objects, enabling the generation of multiple visual matching QA pairs. 
We directly utilize existing mask annotations to refer to objects and form multiple candidate options, while leveraging existing matching annotations to construct the answers. 
For image pairs lacking mask annotations, we employ SAM~\cite{kirillov2023segment} to automatically generate the corresponding mask annotations.
In cases where image pairs lack matching annotations, we utilize the Re-identification method~\cite{li2024masa} to obtain the matching relationships between objects.
Thus, we generate 220K QA pairs in total, the entire process is shown in left side of Fig.~\ref{fig:sft_data}.

\noindent\textbf{Reason Generation.} Multiple-choice training data can hardly provide text supervision for MLLMs. 
Inspired by chain-of-thought~\cite{wei2022chain}, we append reasoning and explanation for each multiple-choice question. 
For this purpose, we design a pipeline to prompt MLLMs with mask annotations and matching annotations to generate reasons automatically. 
Although our experiments indicate that existing MLLMs exhibit poor visual matching capabilities, within our pipeline, MLLMs are not required to perform visual matching themselves. 
Instead, they only need to summarize visual cues that are beneficial for visual matching based on the provided annotations.
As shown in the middle of Fig.~\ref{fig:sft_data}, first, we prompt the stronger MLLM InternVL2-76B~\cite{chen2024internvl} to annotate basic information for all query and candidate objects, including color, size, position, posture, etc. 
Then, we give both the answer (which two objects are the same) and the objects' basic information as conditions and prompt InternVL2-76B to generate corresponding reasons. 
Finally, we obtain 220K matching QA pairs with reasons.

\noindent\textbf{Match VQA Formulation.} As illustrated on the right side of Fig.~\ref{fig:sft_data}, we organize the multiple-choice QA pairs and reasoning texts generated in Step 1 and Step 2 into a two-turn dialogue. The first turn requires the model to make a selection, whereas the second turn requires the model to provide a reasoning for its chosen answer. Ultimately, we obtain a dataset comprising 220k multi-turn dialogue samples.

\subsection{MMVM Benchmark}
\label{sec:evaluate_data}

To evaluate the visual matching capabilities of MLLMs, we also collect image pairs from internet video platforms and the validation splits of existing video datasets. 
These pairs are manually annotated with mask annotations and matching annotations by three experts.
We specifically select challenging examples to form our MMVM benchmark, which comprise a total of 1,510 examples.

\noindent\textbf{Example Format.} 
As shown in Fig.~\ref{fig:benchmark}, we use text prompts or visual prompts to specify objects. 
Considering that most MLLMs cannot understand additional visual prompts, we overlay the visual prompts onto the images using highlight contours of different colors and a number tag. 
Each example consists of image pairs (more than two images), a question, and options. 
The MLLM must select the correct answer from the given options based on the question and image pairs.

\noindent\textbf{Benchmark Statistics.} 
Our MMVM benchmark comprise a total of 1,510 examples. Among these, 790 examples are derived from 15 video segmentation, tracking, and multi-view matching datasets (including OVIS~\cite{qi2022occluded}, Youtube-VIS 2021~\cite{yang2019video}, LVVIS~\cite{wang2024ov}, MOSE~\cite{ding2023mose}, BDD100k~\cite{yu2020bdd100k}, BURST~\cite{athar2023burst}, SportMOT~\cite{cui2023sportsmot}, VideoCube~\cite{Hu2023videocube}, MultiviewX~\cite{hou2020multiview}, VastTrack~\cite{peng2024vasttrack}, MOT17~\cite{milan2016mot17}, LaSOT~\cite{fan2019lasot}, VIPSeg~\cite{miao2022vipseg}, WildTrack~\cite{chavdarova2018wildtrack}, and OxUvA~\cite{valmadre2018long}). 
To mitigate potential overlap with MMVM training data (Sec.~\ref{sec:train_data}), we exclusively selected image pairs from the validation splits of these datasets and re-annotated them manually. To further augment the diversity and complexity of the MMVM benchmark, we manually gathered 720 videos from a variety of internet video platforms. The diversity of these 1,510 examples enables us to evaluate the ability of MLLMs to discern and comprehend multiple matching cues. As illustrated in Fig.~\ref{fig:mmvm_statistics}, we enumerate eight types of matching cues, including: 1) Color (CL), 2) Shape or posture (SP), 3) Textual or LOGO markers (TM), 4) Size (SZ), 5) Relative position in the scene (RP), 6) Object orientation and movement (OO), 7) Binding relationship with other objects (BR), and 8) Object Markers (OM). 
The examples of these matching cues are shown in Fig.~\ref{fig:benchmark}. Each example may exhibit multiple matching cues, but we annotate only the most salient ones. CL and RP are the most prevalent cues. Each example includes multiple candidate options, as depicted in Fig.~\ref{fig:mmvm_statistics}. The minimum, maximum, and average number of choices per example are 2, 37, and 10, respectively.

\begin{figure*}[!t]
\centering
\vspace{-2mm}\includegraphics[width=0.85\textwidth]{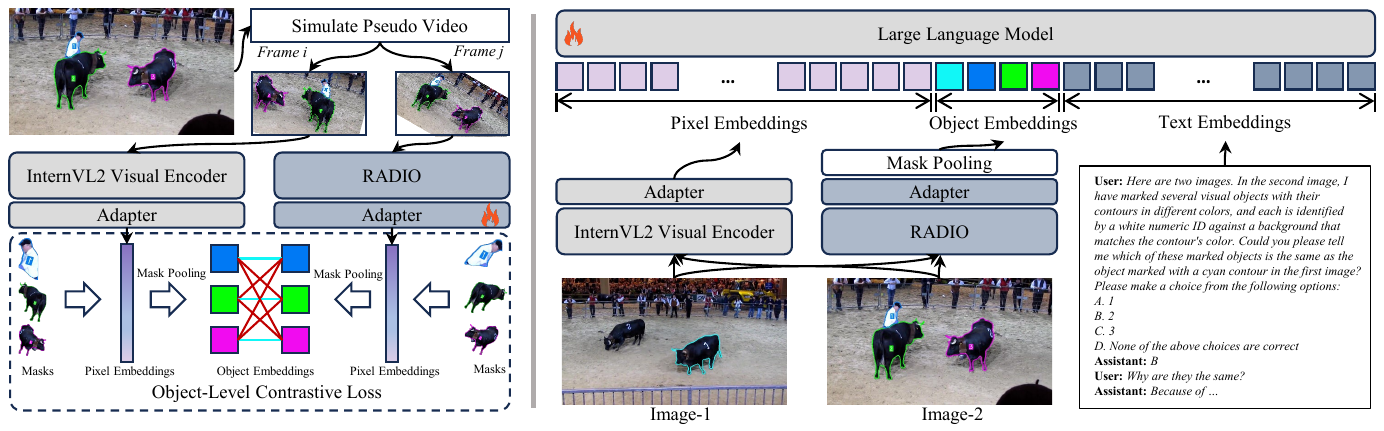}
\vspace{-2mm}\caption{\small The overview of CoLVA. The left side shows how we use object-level contrastive loss to train the RADIO adapter to simultaneously obtain discriminative features and align the RADIO feature space with MLLM's feature space. The right side shows how we integrate the learned contrastive visual tokens into the MLLMs. We directly concatenate the learned contrastive visual tokens with the origin visual tokens output from the MLLM's visual encoder and feed them into the MLLM's LLM for answer generation.}\vspace{-2mm}
\label{fig:model}
\end{figure*}

\noindent\textbf{Evaluation Metric.} Following previous works~\cite{fu2024mme,balanced_vqa_v2}, we adopt accuracy as the main evaluation metric. In addition to calculating an overall accuracy, we also separately assess accuracy for eight distinct matching cues. 
%

\section{Method}
\label{sec:method}

\subsection{Analysis of Current MLLMs' Shortcomings}
\label{sec:shortcomming}

We have evaluated multiple SOTA open-source and proprietary MLLMs on the MMVM benchmark. However, the results reveal a notable observation: none of the open-source MLLMs achieve an overall accuracy exceeding 50\% (Tab.~\ref{tab:match_bench_result}).
This phenomenon suggests significant deficiencies in current MLLMs' performance on the visual matching task.
We argue that two main factors cause this phenomenon: \textbf{1)} \textit{Although current MLLMs possess a certain capability to recognize objects' appearances and positions, they lack the corresponding data to teach them how to utilize this basic knowledge and these abilities for visual matching, even in a simple sense.} 
This hypothesis is supported by two observations: First, in the annotation pipeline of the MMVM dataset, InternVL2 can accurately identify the basic information of query objects, yet it achieves a notably low score on the MMVM benchmark, as illustrated in Tab.~\ref{tab:match_bench_result}. 
Second, when we fine-tune InternVL2 using our MMVM dataset enriched with matching reasoning, its performance improves significantly (+14.76\%).
\textbf{2)} \textit{Current MLLMs rely on CLIP models to understand images and cannot comprehend fine-grained and discriminative visual features, which are essential for visual matching since candidate objects often share extremely similar semantic information.} 
As illustrated in Fig.~\ref{fig:pca}, we conduct a PCA visualization analysis on the object embeddings learned by InternVL2. The results show that the matched target (represented by a red dot) and other candidate objects (represented by blue dots) are clustered together, while being distant from the query object (represented by a red star). This clustering pattern makes it challenging for MLLMs to distinguish the correct object.

\newcommand{\Hmat}{{\bf H}}
\newcommand{\Omat}{{\bf O}}
\newcommand{\Wmat}[0]{{{\bf W}}}
\newcommand{\Xmat}[0]{{{\bf X}}}
\newcommand{\Zmat}{{\bf Z}}
\newcommand{\thetav}{\boldsymbol{\theta}}

\vspace{-1mm}\subsection{CoLVA}\vspace{-1mm}
\label{sec:method_colva}

To address the shortcomings summarized in Sec.~\ref{sec:shortcomming}, we propose a novel Object-level Contrastive Learning (OCL) strategy and introduce a fine-grained vision encoder to provide the discriminative and fine-grained visual features, thereby improving the MLLM's visual matching performance. 
Additionally, we propose an instruction augmentation strategy to facilitate MLLM training.
These two novel technical designs will be detailed in the following.

\noindent\textbf{Baseline.}
Due to its strong single and multi-image QA performance, we select the SOTA MLLM InternVL2~\cite{chen2024internvl} as our baseline. 
We construct a strong baseline by fine-tuning InternVL2 using a combination of LLaVA SFT data~\cite{liu2024llava1.5} and our MMVM data.

\noindent\textbf{Object-Level Contrastive Learning.}
Inspired by the success of contrastive learning in visual tracking~\cite{he2020momentum, chen2020improved, oquab2023dinov2}, tracking~\cite{yan2022towards, yan2023universal}, and video segmentation~\cite{wu2022defense, zhang2023dvis++, li2023tube, ying2023ctvis,li2024masa}, we introduce a novel object-level contrastive learning (OCL) strategy to help MLLM learn more discriminative features for better visual matching.
Unlike previous contrastive learning approaches that learning features using shared tracking heads, our method conducts contrastive learning between two distinct vision encoders: the MLLM visual encoder and an additional visual expert, as illustrated on the left side of Fig.~\ref{fig:model}. 
This design allows the visual expert to learn discriminative features within the semantic space of the MLLM.
On the one hand, it preserves fine-grained visual features, while on the other hand, it achieves modality alignment through contrastive learning.
We employ the OCL strategy during the pre-training phase. 
As shown in Tab.~\ref{tab:align_method_ablation}, the OCL strategy outperforms other pre-training methods.

Firstly, we obtain the object-level representations using masked average pooling based on the image feature. 
Then, the object-level contrastive loss is conducted on the object-level representations:
\begin{equation}
\mathcal{L} = \frac{exp(\Omat \cdot \Omat^+)}{exp(\Omat \cdot \Omat^+) + \sum_{\Omat^-}exp(\Omat \cdot \Omat^-)},
\label{eq:contra_loss}
\end{equation}
where $\Omat$ denotes the object-level representation of the query object. $\Omat^+$ and $\Omat^-$ denote the representations of positive and negative candidate objects, respectively.

\noindent\textbf{Fine-grained Vision Expert.} 
We find that directly applying OCL on MLLM's vision backbone only achieves limited improvement (34.05 vs. 32.38, as shown in Tab.~\ref{tab:main_ablation}). It is because MLLM's CLIP-style backbone lacks fine-grained visual features. 
Inspired by~\cite{shi2024eagle,tong2024eyes}, we incorporate an additional fine-grained vision expert, RADIO~\cite{ranzinger2024radio}, into the MLLM to provide more powerful visual representations. 
RADIO is distilled from the SAM~\cite{kirillov2023segment} encoder, DINOv2~\cite{oquab2023dinov2}, CLIP~\cite{radford2021learning}, and other vision foundation models, thus possessing comprehensive capabilities such as fine-grained visual features and good image-text alignment ability.
Due to the significant gaps between RADIO's and MLLM's feature spaces, we introduce an additional pre-training stage to align them, and OCL can be easily integrated into this process. 
As depicted on the left side of Fig.~\ref{fig:model}, we incorporate RADIO into the MLLM. 
OCL is used in the pre-training stage to simultaneously obtain discriminative features and align the RADIO feature space with MLLM's feature space. 
For an input image pair, one image is fed into the MLLM's original visual encoder, while another is input into the RADIO. 
We then apply OCL (details in Eq.~\ref{eq:contra_loss}) on all objects' representations. 
Due to the limited image pairs with segmentation annotations, we simulate many pseudo-video data with masklets based on image segmentation datasets following~\cite{zhang2023dvis++}.

In the pre-training stage, we freeze the InternVL2's visual encoder, the InternVL2 adapter, and RADIO, focusing solely on training the RADIO adapter. 
After the pre-training stage, RADIO's feature space is aligned with the MLLM's. 
The MLLM can perform the SFT stage the same as the previous methods~\cite{Liu2023llava,chen2024internvl}.

\noindent\textbf{Instruction Augmentation.}
In the SFT stage, we use highlighted contours to mark the query and candidate objects and draw corresponding ID tags. 
The instruction data can be summarized in the format:

\textit{\small{``\textless Edited\_IMGs\textgreater\textbackslash n\textless SYSTEM\textgreater \textless Question\_Answer\textgreater"}}

\noindent where \textit{\small{\textless SYSTEM\textgreater}} is: \textit{``Here are two images. In the second image, I have marked several visual objects with their contours in different colors, and each is identified by a white ID against a background that matches the contour's color."}

This instruction format allows the MMVM data to be seamlessly compatible with InternVL2~\cite{chen2024internvl} for direct training. However, it still has some drawbacks. 1) Editing the image may disrupt the original object information, especially for small objects. 2) Since the tags are used indirectly to refer to objects, gradients cannot be directly backpropagated to the corresponding image features.
To address these problems, we designed a new instruction format to support direct use of object-level representations to refer to objects:

\noindent\textit{\footnotesize{``\textless Edited\_IMGs\textgreater\textbackslash n\textless SYSTEM\textgreater \textless Object\_Info\textgreater \textless Question\_Answer\textgreater"}}

\noindent where \textit{\small{\textless Object\_Info\textgreater}} is: \textit{\small{``object-1: \textless Obj 1\textgreater, object-2: \textless Obj 2\textgreater, ..., object-n: \textless Obj n\textgreater"}}, with \textit{\small{``\textless Obj 1\textgreater"}} to \textit{\small{``\textless Obj n\textgreater"}} replaced by the respective object-level representations. 
This instruction format allows gradients to be directly backpropagated through the object-level representations to the corresponding image features, enabling the MLLM to learn the required discriminative and fine-grained features more effectively. We randomly use these two instruction formats to organize the data, which we refer to as instruction augmentation. 

%% file: ICCV/latex/4_exp.tex
\section{Experiments}
\label{sec:exp}

\begin{table*}[t!]
    \centering
    \caption{MMVM Benchmark Results. Given that CL and RP are the two primary matching cues, we report the overall accuracy, CL accuracy, and RP accuracy under four different settings: with 4, 8, 12, and all candidate options. The overall accuracy is computed across all 1,510 evaluation samples. CL accuracy and RP accuracy are calculated separately on their respective samples. The full terms of the matching type abbreviations can be found in Sec.~\ref{sec:evaluate_data}. For MLLMs that only support single-image input, we simply concatenate all the images vertically into a single image and then provide it as input.}\vspace{-2mm}
    \label{tab:match_bench_result}
    \resizebox{0.85\textwidth}{!}{
    \begin{tabular}{l | l | c c c | c c c | c c c | c c c}
    \toprule[1.5pt]
        Model Size & Method & Overall & CL & RP & Overall-12 & CL-12 & RP-12 & Overall-8 & CL-8 & RP-8 & Overall-4 & CL-4 & RP-4 \\
        \midrule[1pt]
        \multirow{3}{*}{$\sim$4B} & Qwen2-VL-2B-Instruct~\cite{wang2024qwen2} & 15.69 & 13.42 & 9.57 & 16.82 & 14.57 & 11.71 & 19.93 & 17.68 & 16.28 & 32.35 & 28.97 & 30.86 \\
        ~ & DeepSeek-VL-1.3B~\cite{lu2024deepseek} & 16.82 & 12.60 & 10.43 & 17.46 & 13.97 & 10.70 & 21.77 & 18.36 & 19.18 & 28.21 & 25.96 & 23.58 \\
        ~ & InternVL2-4B~\cite{chen2024internvl} & 17.62 & 14.73 & 10.28 & 18.34 & 18.17 & 14.86 & 20.73 & 20.78 & 18.71 & 35.76 & 35.84 & 35.00 \\
        \midrule[1pt]
        \multirow{4}{*}{4B$\sim$13B} & DeepSeek-VL-7b~\cite{lu2024deepseek} & 17.68 & 14.24 & 10.00 & 19.22 & 14.79 & 11.98 & 23.04 & 18.70 & 16.29 & 27.01 & 24.13 & 20.56 \\
        ~ & LLaVA-Next-Interleave-7B~\cite{li2024llavanextinter} & 19.34 & 15.88 & 10.71 & 21.03 & 17.75 & 12.86 & 25.01 & 19.53 & 16.64 & 28.47 & 26.04 & 20.36 \\
        ~ & LLaVA-OneVision-ov-7B~\cite{li2024llavaonevision} & 20.92 & 16.69 & 14.28 & 22.01 & 17.96 & 15.47 & 25.85 & 21.30 & 19.67 & 29.87 & 25.46 & 23.33 \\
        ~ & Qwen2-VL-7B-Instruct~\cite{wang2024qwen2} & 27.48 & 24.87 & 17.85 & 28.34 & 25.51 & 18.97 & 30.04 & 26.98 & 22.11 & 36.75 & 33.21 & 27.30 \\
        \midrule[1pt]
        \multirow{3}{*}{13B$\sim$40B} & LLaVA-Next-34B~\cite{liu2024llavanext} & 15.03 & 11.29 & 8.71 & 15.86 & 12.33 & 9.22 & 19.79 & 16.84 & 11.79 & 24.97 & 21.46 & 17.83 \\
        ~ & VILA1.5-40B~\cite{lin2024vila} & 15.36 & 14.73 & 5.00 & 16.95 & 15.60 & 6.45 & 20.85 & 18.78 & 12.01 & 26.12 & 24.37 & 17.33 \\
        ~ & InternVL2-40B~\cite{chen2024internvl} & 26.03 & 24.88 & 16.86 & 27.99 & 26.13 & 19.46 & 32.63 & 30.70 & 24.97 & 37.35 & 34.93 & 30.05 \\
        \midrule[1pt]
        \multirow{4}{*}{40B$\sim$} & InternVL2-76B~\cite{chen2024internvl} & 25.83 & 24.06 & 19.28 & 27.56 & 26.13 & 21.99 & 32.76 & 30.07 & 26.94 & 36.17 & 34.96 & 30.54 \\
        ~ & LLaVA-OneVision-ov-72B~\cite{li2024llavaonevision} & 29.34 & 28.48 & 21.14 & 29.89 & 29.31 & 23.14 & 32.55 & 33.03 & 27.34 & 37.43 & 35.20 & 30.34 \\
        ~ & InternVL2.5-78B~\cite{chen2024internvl2_5} & 36.42 & 35.02 & 25.86 & 39.01 & 35.72 & 33.40 & 42.11 & 39.31 & 34.79 & 45.89 & 43.41 & 37.57 \\
        ~ & Qwen2-VL-72B-Instruct~\cite{wang2024qwen2} & 38.08 & 37.64 & 32.28 & 39.77 & 35.00 & 31.94 & 42.31 & 39.83 & 35.62 & 47.68 & 45.23 & 39.23 \\
        \midrule[1pt]
        \multirow{3}{*}{Unkown} & Claude3-5V-Sonnet & 40.20 & 34.21 & 34.86 & 41.75 & 33.98 & 37.01 & 45.77 & 38.02 & 41.39 & 51.35 & 43.79 & 48.63 \\
        ~ & GeminiPro1-5 & 40.73 & 36.00 & 35.14 & 43.01 & 39.27 & 36.98 & 46.66 & 41.97 & 38.07 & 52.62 & 50.30 & 45.13 \\
        ~ & GPT4o-20240806 & 42.65 & 39.28 & 32.28 & 44.71 & 43.63 & 39.65 & 49.30 & 45.37 & 42.11 & 56.76 & 53.24 & 47.57 \\
        
        \midrule[1pt]
                4B & CoLVA-InternVL2-4B (Ours) & 49.80 & 42.72 & 44.86 & 51.06 & 44.19 & 46.86 & 53.38 & 46.48 & 49.71 & 59.47 & 51.72 & 57.86 \\
    \bottomrule[1.5pt]
    \end{tabular}
    }\vspace{-4mm}
\end{table*}

\begin{table}[h]
    \centering
    \caption{The impact of CoLVA on the single-image VQA capabilities of MLLMs.}
    \label{tab:colva_on_single_image_bench}
    \resizebox{0.45\textwidth}{!}{
    \begin{tabular}{c c|c c c c c c c}
    \toprule[1.5pt]
         \multirow{2}{*}{MLLM} & \multirow{2}{*}{CoLVA} & MMBench DEV & MME & MME & MMStar & MMMU Val & POPE & BLINK \\
         ~ & ~ & Overall & Perception & Reasoning & Overall & Overall & Overall & Overall\\
         \midrule[1pt]
         \multirow{2}{*}{InternVL2-4B~\cite{chen2024internvl}} & $\times$ & 77.40 & 1536.14 & 533.93 & 54.40 & 47.56 & 84.52 & 45.76\\
         ~ & \checkmark & 77.32 & 1552.82 & 549.64 & 53.47 & 44.11 & 86.11 & 47.24\\
    \bottomrule[1.5pt]
    \end{tabular}
    }
\end{table}

\begin{table}[h]
    \centering
    \caption{The impact of CoLVA on the multi-image VQA capabilities of MLLMs.}
    \label{tab:colva_on_multi_image_bench}
    \resizebox{0.45\textwidth}{!}{
    \begin{tabular}{c c|c c c c}
    \toprule[1.5pt]
         \multirow{2}{*}{MLLM} & CoLVA & NaturalBench & NaturalBench & NaturalBench & VideoRefer-Bench \\
         ~ & ~ & Q\_Acc & I\_Acc & G\_Acc & Average \\
         \midrule[1pt]
         \multirow{2}{*}{InternVL2-4B~\cite{chen2024internvl}} & $\times$ & 44.71 & 48.63 & 19.52 & 60.91\\
         ~ & \checkmark & 47.89 & 52.16 & 20.84 & 62.94\\
    \bottomrule[1.5pt]
    \end{tabular}
    }
\end{table}

\noindent
\textbf{Baselines and Datasets.} We use the pre-trained InternVL2 4B~\cite{chen2024internvl} as the baseline. During the SFT phase, we utilize the LLaVA SFT data~\cite{liu2024llava1.5} and our MMVM dataset. The LLaVA SFT data comprises approximately 665k conversation entries, and our MMVM data includes around 220k. \textit{Please note that all ablation experiments were conducted using 30\% of these data.}

\noindent
\textbf{Implementation Details.} Our model comprises three components: a pre-trained MLLM InternVL2-4B~\cite{chen2024internvl}, a fine-grained vision expert RADIO~\cite{ranzinger2024radio}, and a RADIO adapter.
We adopt Xtuner~\cite{2023xtuner} codebase to implement our method.
Please refer to the appendix for the details.

%


\subsection{Main Results}
\label{sec:exp_main_results}

\noindent
\textbf{Results on MMVM benchmark.} As shown in Tab.~\ref{tab:match_bench_result}, we report the average accuracy of multiple open-source MLLMs of varying sizes, three proprietary MLLMs, and our method on the MMVM Benchmark. 
In the MMVM benchmark, none of the open-source or proprietary MLLMs achieved an overall accuracy exceeding 50\% under the setting of all choice options.
Compared to relative position in the scene, these MLLMs demonstrate a stronger capability in perceiving and utilizing color as a matching cue, as evidenced by the CL accuracy consistently surpassing the RP accuracy across almost all settings.
By introducing object-level contrastive learning, fine-grained vision expert, and instruction augmentation, our method achieved significant performance improvements, reaching state-of-the-art performance and surpassing the previous highest accuracy obtained by GPT4o. 
Among all open-source MLLMs, InternVL2~\cite{chen2024internvl} achieved the highest accuracy in the sub-4B and 13B$\sim$40B tiers, while Qwen2-VL~\cite{wang2024qwen2} excelled in the 4B$\sim$13B and above-40B tiers. 
Overall, Qwen2-VL-72B~\cite{wang2024qwen2} achieved the highest accuracy among all open-source MLLMs, approaching the accuracy of the proprietary GPT4o (38.08 vs. 42.65).

\noindent
\textbf{Results on common VQA benchmarks.}
To investigate whether CoLVA adversely affects the inherent general visual question answering (VQA) capabilities of MLLMs, we conducted tests across six relevant benchmarks: MMBench~\cite{liu2025mmbench}, MME~\cite{fu2024mme}, MMStar~\cite{chen2024mmstar}, MMMU~\cite{yue2024mmmu}, POPE~\cite{li2023pope}, BLINK~\cite{fu2025blink}, NaturalBench~\cite{li2024naturalbench}, and VideoRefer-Bench~\cite{yuan2024videorefer}. The results are presented in Tab.~\ref{tab:colva_on_single_image_bench} and Tab.~\ref{tab:colva_on_multi_image_bench}. We used InternVL2-4B as the baseline and integrated our CoLVA into this framework. 
The results indicate that the negative impact of CoLVA on the general VQA capabilities of MLLMs is minimal. 
In fact, it even shows positive effects on the MME, POPE, BLINK, NaturalBench, and VideoRefer-Bench benchmarks. 
Therefore, our CoLVA does not compromise the original general VQA capabilities of MLLMs and can be a good supplement to current mainstream VQA datasets.

\subsection{Ablation Study and Analysis}
\label{sec:exp_ablation_study}

\noindent
\textbf{The Effectiveness of Data and Methods.}
As shown in Tab.~\ref{tab:main_ablation}, we used InternVL2-4B as our base model, achieving an overall accuracy of 17.62\% on our MMVM benchmark. By fine-tuning InternVL2-4B with LLaVA SFT data~\cite{liu2024llava1.5} and our MMVM data, we observed a significant increase in overall accuracy (+14.76\%), validating the effectiveness of our MMVM SFT data. We adopted this fine-tuned InternVL2-4B as a strong baseline and integrated our methods, which include object-level contrastive learning, fine-grained vision expert, and instruction augmentation. 
By incorporating the fine-grained vision expert into the fine-tuned InternVL2 and using object-level contrastive learning to pre-train its adapter, we observed an 8.07\% improvement in overall accuracy. 
We believe that VE provides the basic knowledge necessary for object matching, such as recognizing appearance, position, size, and other attributes. 
OCL offers an appropriate training strategy to enable MLLM to comprehend the knowledge provided by VE. Sufficient object matching data is crucial for teaching the MLLM to utilize this knowledge to perform object matching.
Due to the substantial gap between the feature space of the fine-grained vision expert and that of the MLLM, directly using the visual features from the fine-grained vision expert did not yield any significant impact (32.25 vs. 32.38). 
Notably, directly applying object-level contrastive learning to the visual encoder of InternVL2 resulted in only a limited improvement in overall accuracy (34.05 vs. 32.38), as the CLIP-style vision backbone lacks fine-grained visual features. 
Further augmentation of instructions led to an additional accuracy gain of 5.38\%. 


\noindent
\textbf{The Effectiveness of Object-level Contrastive Learning.}
Our method employs object-level contrastive learning to pre-train the RADIO adapter. As shown in Tab.~\ref{tab:align_method_ablation}, compared to other standard methods that use image-text pairs (Image-Text) or region-text pairs (Region-Text) to pre-train the adapter by applying autoregressive training objective, our method (Region-Region) demonstrates significant advantages (40.45 vs. 33.64, or 40.45 vs. 30.93). 


\noindent
\textbf{The Alternatives of RADIO.}
Our method still works well for vision self-supervised learning models. 
In particular, we replaced RADIO with DINOv2~\cite{oquab2023dinov2} and ConvNext-L~\cite{woo2023convnext}. 
As shown in Tab.~\ref{tab:alter_radio}, our method still proves effective for vision-only SSL models, with a significant improvement in accuracy (40.34 vs 32.38).
However, there is a gap between RADIO and DINOv2.
This means both a semantic and spatial-aware visual expert is needed to achieve better results.
CoLVA with ConvNext CLIP demonstrates a superior understanding of text markers (TM) compared to DINOv2 and RADIO but exhibits worse overall performance.



\begin{table}[t]
    \centering
    \caption{The effectiveness of our methods and MMVM data. Data denotes using the combination of MMVM data and LLaVA SFT data. OCL denotes object-level contrastive learning. VE denotes fine-grained vision expert. IA denotes instruction augmentation.}
    \label{tab:main_ablation}
    \resizebox{0.30\textwidth}{!}{
    \begin{tabular}{cccc| cc}
    \toprule[1.5pt]
        Data & OCL & VE & IA & OA & $\Delta$\\
        \midrule[1pt]
         ~ & ~ & ~ & ~ & 17.62 & -\\
         \checkmark & ~ & ~ & ~ & 32.38 &\textcolor[rgb]{0, 1, 0}{+14.76} \\
         \checkmark & \checkmark & ~ & ~ & 34.05 &\textcolor[rgb]{0,1,0}{+1.67} \\
         \checkmark & ~ & \checkmark & ~ & 32.25 &\textcolor[rgb]{1,0,0}{-0.13}  \\
         \checkmark & \checkmark & \checkmark & ~ & 40.45 &\textcolor[rgb]{0, 1, 0}{+8.07}\\
        \checkmark & \checkmark & \checkmark & \checkmark & 45.83 &\textcolor[rgb]{0, 1, 0}{+5.38}\\
    \bottomrule[1.5pt]
    \end{tabular}
    }\vspace{-4mm}
\end{table}

\begin{table}[t]
    \centering
    \caption{The effectiveness of object-level contrastive loss. Image-Text/Region-Text means using image-text/region-text pairs for pre-training. Region-Region means using contrastive loss for pre-training.}\vspace{-2mm}
    \label{tab:align_method_ablation}
    \resizebox{0.48\textwidth}{!}{
    \begin{tabular}{c | c c c c c}
    \toprule[1.5pt]
        Metric & Baseline & No Alignment & Image-Text & Region-Text & Region-Region \\
        \midrule[1pt]
        Overall Acc. & 32.38 & 32.25  & 33.64 & 30.93 & 40.45 \\
    \bottomrule[1.5pt]
    \end{tabular}
    }\vspace{-4mm}
\end{table}

\begin{table}[t]
    \centering
    \caption{The alternatives of RADIO. The baseline is without any fine-grained vision expert.}\vspace{-2mm}
    \label{tab:alter_radio}
    \resizebox{0.48\textwidth}{!}{
    \begin{tabular}{c | c c c c c c c c c}
    \toprule[1.5pt]
        ~ & Overall & CL & SP & TM & SZ & RP & OO & BR & OM \\
        \midrule[1pt]
        Baseline & 32.38 & 25.04 & 24.14 & 32.71 & 74.03 & 19.00 & 35.20 & 43.18 & 36.57 \\
        RADIO~\cite{ranzinger2024radio} & 45.83 & 38.30 & 31.03 & 41.12 & 76.62 & 41.71 & 51.20 & 39.77 & 46.76 \\
        DINOv2~\cite{oquab2023dinov2} & 40.34 & 33.72 & 44.83 & 42.06 & 64.94 & 32.28 & 36.00 & 35.80 & 39.81 \\
        ConvNext-L~\cite{woo2023convnext} & 39.80 & 31.59 & 34.48 & 42.99 & 77.92 & 26.57 & 48.80 & 44.32 & 44.44 \\
    \bottomrule[1.5pt]
    \end{tabular}\vspace{-5mm}
    }
\end{table}




\subsection{Generalization study of CoLVA}
\label{sec:gene_colva}


\begin{table}[h]
    \centering
    \caption{The effectiveness of CoLVA on more MLLMs. OA denotes the overall accuracy. 
    }
    \label{tab:colva_on_more_mllms}
    \resizebox{0.3\textwidth}{!}{
    \begin{tabular}{c c|c c c }
    \toprule[1.5pt]
         MLLM & CoLVA & OA & CL & RP \\
         \midrule[1pt]
         \multirow{2}{*}{InternVL2-4B~\cite{chen2024internvl}} & $\times$ & 17.62 & 14.73 & 10.28 \\
         ~ & \checkmark & 45.83 & 38.30 & 41.71 \\
         \midrule[1pt]
         \multirow{2}{*}{Qwen2VL-2B~\cite{wang2024qwen2}} & $\times$ & 15.69 & 13.42 & 9.57 \\
         ~ & \checkmark & 47.48 & 40.92 & 50.57 \\
         \midrule[1pt]
         \multirow{2}{*}{LLaVA1.5-7B~\cite{liu2023improvedllava}} & $\times$ & 14.64 & 12.44 & 8.00 \\
         ~ & \checkmark & 36.56 & 29.13 & 26.14 \\
    \bottomrule[1.5pt]
    \end{tabular}
    }
    \vspace{-4mm}
\end{table}


To validate the generalization of CoLVA on different MLLMs, we integrate CoLVA into three distinct MLLMs: InternVL2-4B~\cite{chen2024internvl}, Qwen2VL-2B~\cite{wang2024qwen2}, and LLaVA1.5-7B~\cite{liu2024llava1.5}. 
Their performance on the MMVM benchmark is presented in Tab.~\ref{tab:colva_on_more_mllms}. 
The results demonstrate that CoLVA significantly improves the fine-grained visual matching capabilities across all three MLLMs. 
Notably, LLaVA1.5-7B, which has not undergone multi-image training, exhibited the smallest accuracy improvement after integrating CoLVA. 
In contrast, both InternVL2-4B and Qwen2VL-2B, having been trained with multiple images, showed substantial accuracy improvements on the MMVM benchmark with our CoLVA integration. 
%


%% file: ICCV/latex/5_conclusion.tex
\section{Conclusion}
\label{sec:conclusion}

This paper presents the MMVM benchmark, the first corresponding fine-grained visual correspondence evaluation benchmark for current MLLMs. 
The results demonstrate all MLLMs perform poorly, with none achieving accuracy above 50\% under the setting of all candidate options, including GPT-4o. 
To address the significant weakness of current MLLMs in visual correspondence, we design an automatic annotation pipeline to generate a 220K visual matching SFT dataset with reasoning. 
Furthermore, we propose CoLVA through two novel designs: combining object-level contrastive learning with RADIO to obtain descriptive visual features and an instruction augmentation strategy. 
Experiments demonstrate that our novel designs improve base MLLM by 13.45 OA. 
Benefiting from our SFT data and the novel designs, our proposed CoLVA-InternVL2-4B achieves 49.80 OA on the MMVM benchmark, surpassing the baseline InternVL2-4B with a 32.18 OA performance improvement.


%% file: ICCV/latex/6_appendix.tex
\appendix


\begin{table*}[!ht]
    \centering
    \caption{The effectiveness of our methods and MMVM data 
with detailed results. Data denotes using the combination of MMVM data and LLaVA SFT data. OCL denotes object-level contrastive learning. VE denotes fine-grained vision expert. IA denotes instruction augmentation. OA denotes the overall accuracy. 
}
    \label{tab:main_ablation_detailed}
    \resizebox{0.80\textwidth}{!}{
    \begin{tabular}{cccc| ccccccccc } 
    \toprule[1.5pt]
        Data & OCL & VE & IA & OA & CL & SP & TM & SZ & RP & OO & BR & OM\\
        \midrule[1pt]
         ~ & ~ & ~ & ~ & 17.62 & 14.73 & 34.48 & 17.76 & 15.58 & 10.28 & 24.00 & 31.25 & 21.30\\
         \checkmark & ~ & ~ & ~ & 32.38 & 25.04 & 24.14 & 32.71 & 74.03 & 19.00 & 35.20 & 43.18 & 36.57 \\
         \checkmark & \checkmark & ~ & ~ & 34.05 & 25.78 & 26.77 & 31.97 & 75.01 & 22.32 & 35.29 & 42.98 & 37.51\\
         \checkmark & ~ & \checkmark & ~ & 32.25 & 24.22 & 27.59 & 31.78 & 68.83 & 19.14 & 35.20 & 40.34 & 39.35  \\
         \checkmark & \checkmark & \checkmark & ~ & 40.45 & 33.72 & 44.85 & 39.37 & 75.33 & 30.00 & 48.00 & 38.65 & 44.78\\
        \checkmark & \checkmark & \checkmark & \checkmark & 45.83 & 38.30 & 31.03 & 41.12 & 76.62 & 41.71 & 51.20 & 39.77 & 46.76\\
    \bottomrule[1.5pt]
    \end{tabular}
    }
\end{table*}

\begin{table}[h]
    \centering
    \caption{The impact of Qwen2VL-CoLVA on general benchmarks.}
    \label{tab:qwen_colva}
    \resizebox{0.44\textwidth}{!}{
    \begin{tabular}{c c|c c c c }
    \toprule[1.5pt]
         \multirow{2}{*}{MLLM} & \multirow{2}{*}{CoLVA} & MME & MME & POPE & BLINK \\
         ~ & ~ & perception & reasoning & Overall & Overall \\
         \midrule[1pt]
         \multirow{2}{*}{Qwen2VL-2B} & $\times$ & 1471.10 & 404.64 & 86.83 & 44.50 \\
         ~ & \checkmark & 1540.14 & 418.57 & 88.01 & 46.98 \\
    \bottomrule[1.5pt]
    \end{tabular}
    }
\end{table}

\begin{table}[h]
    \centering
    \caption{The split of MMVM benchmark.}
    \label{tab:split}
    \resizebox{0.44\textwidth}{!}{
     \begin{tabular}[t]{c|c c c}
        \toprule[1pt]
            Method & Total & In-domain split & Out-domain split  \\
            \midrule[0.5pt]
            GPT4o & 42.65 & 46.46 & 38.47 \\
            InternVL2-4B & 17.62 & 21.01 & 13.89 \\
            CoLVA-4B & 49.87 & 57.22 & 41.67 \\
        \bottomrule[1pt]
    \end{tabular}
    }
\end{table}

\section{More Experiment Result}
\label{sec:supp-compare}

\noindent
\textbf{Ablation studies in more detailed results.}
Here, we present the detailed results of the main ablation experiments, as shown in Tab.~\ref{tab:main_ablation_detailed}. The table includes the overall accuracy and accuracy across eight different match types. Our method significantly improves accuracy over a strong baseline (45.83 vs. 32.38) across six match types. The improvement is less pronounced for the size (SZ) match type, where accuracy is approaching saturation (76.62 vs. 74.03).

\noindent
\textbf{CoLVA on the other base model.} 
We combine CoLVA into Qwen2VL and test it on several general benchmarks, as shown in Tab.~\ref{tab:qwen_colva}. CoLVA still works better.

\noindent
\textbf{Analysis on Different Match Types.} From detailed results of Tab.~\ref{tab:more_match_bench_result}, MLLMs work better in matching based on object size (SZ), shape (SP), and textual or LOGO markers (TM). These three types require focusing solely on the object itself, indicating that current MLLMs possess proficient object-level perception and understanding. 
In contrast, MLLMs find it more challenging to match based on object relative position (RP), object orientation and movement (OO), and binding relationships with other objects (BR). 
These require MLLMs to understand the interrelationships between objects and infer information that remains invariant across time and space.

\begin{figure}[t]
\centering
\includegraphics[width=0.50\textwidth]{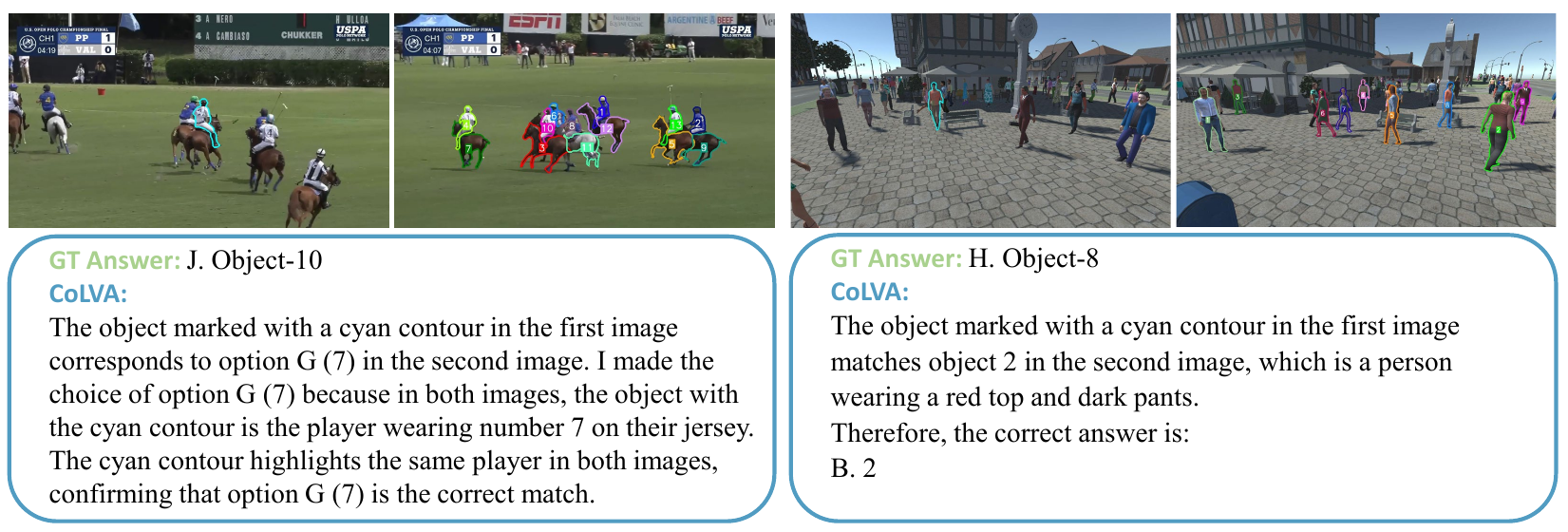}
\vspace{-6mm}\caption{\small The failure cases of CoLVA on MMVM benchmark. CoLVA tends to fail when performing matching in densely populated object scenarios.}\vspace{-5mm}
\label{fig:colva_fail_cases}
\end{figure}

\noindent
\textbf{CoLVA Failure Cases Analysis.}
We have observed that CoLVA tends to fail when performing matching in densely populated object scenarios, as illustrated in Fig.~\ref{fig:colva_fail_cases}. One reason for this is that CoLVA is prone to hallucinations regarding the query object in multi-object, multi-image contexts. For instance, in the left example of Fig.~\ref{fig:colva_fail_cases}, CoLVA correctly identifies the query object as a player. However, in the second image, it mistakenly hallucinates object-7, which is actually a horse, as the matched player. Additionally, in multi-view scenarios, CoLVA is susceptible to incorrectly matching another object based on partial information of the query object from a single viewpoint.

\begin{figure}[t]
\centering
\includegraphics[width=0.50\textwidth]{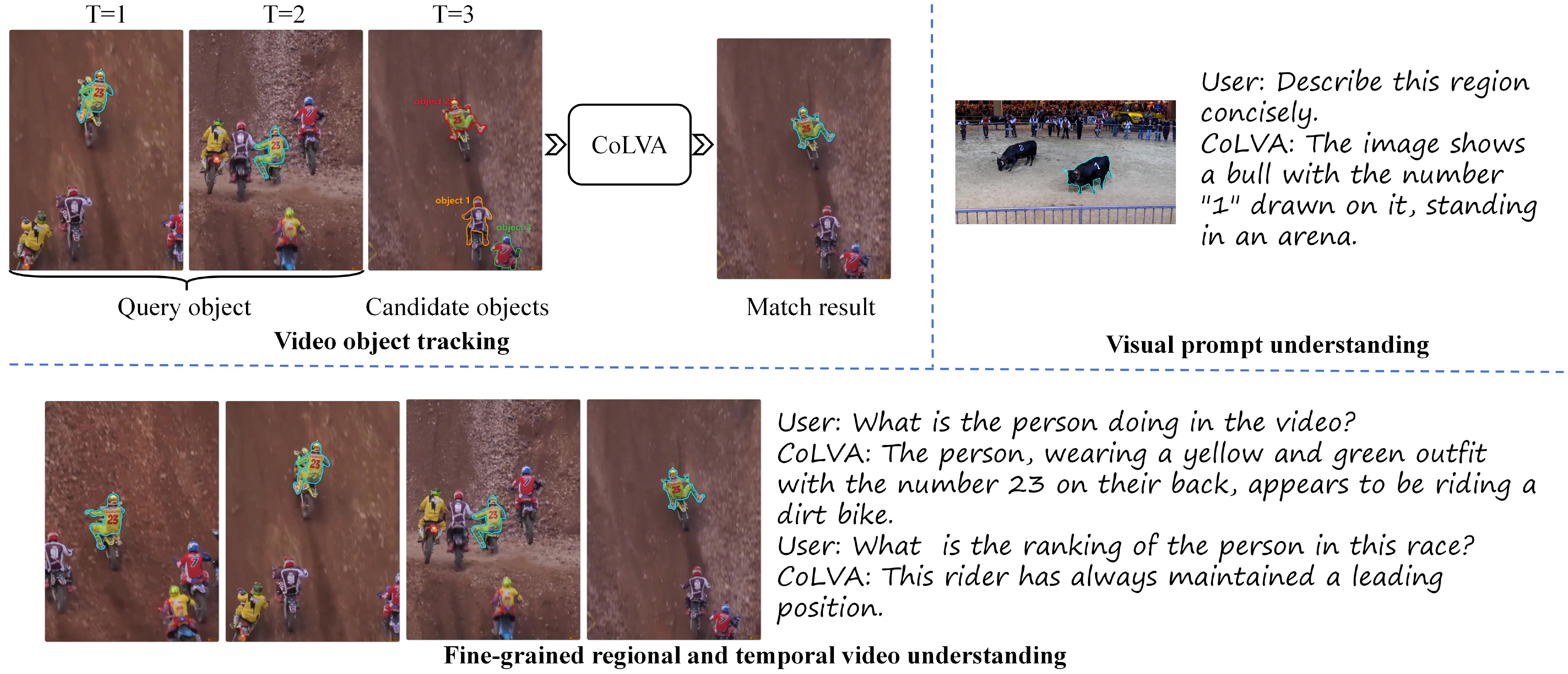}
\vspace{-6mm}\caption{\small Potential real-world applications of CoLVA}\vspace{-5mm}
\label{fig:app}
\end{figure}

\section{More information on MMVM Benchmark}
\label{sec:more_info_mmvm}
The MMVM benchmark is composed of the validation split from the video segmentation datasets (790 samples) and \textbf{manually collected} internet videos (720 samples). 
Additionally, the benchmark is \textbf{not} generated using the automated annotation pipeline employed for the training set, as it only requires matching results without the need for reasoning processes.

We categorize the 790 samples as the in-domain part, and the 720 samples as the out-domain part. Tab.~\ref{tab:split} displays the test results of several methods on these two parts, which revealing that our CoLVA model achieves a significant gain in the out-domain segment (41.67 vs 13.89), thereby demonstrating its robust generalization capability.
%


\section{Potential real-world applications of CoLVA}
\label{sec:app_colva}
Object matching is fundamental to many real-world applications, such as video object tracking, re-identification (ReID), multi-image visual question answering (VQA), and video VQA. Our CoLVA also integrates visual prompt understanding capabilities. In Fig.~\ref{fig:app}, we showcase several real-world applications.

\section{More Implementation Details}
\label{sec:supp-implementation-details}

\noindent
\textbf{More training details.}
Our model comprises three components: a pre-trained MLLM InternVL2-4B~\cite{chen2024internvl}, a fine-grained vision expert RADIO~\cite{ranzinger2024radio}, and a RADIO adapter.
We adopt Xtuner~\cite{2023xtuner} codebase to implement our method.
We maintain the original architecture of both InternVL2-4B and RADIO, while the RADIO adapter is implemented using a two-layer MLP. 
Our training includes two stages: pre-training and supervised fine-tuning (SFT). 
We freeze the MLLM and RADIO during the pre-train stage, focusing solely on training the RADIO Adapter. 
During the SFT stage, we freeze the RADIO, the RADIO adapter, and all components of InternVL2-4B except the LLM. 
The LLM of the MLLM is trained by applying LoRA~\cite{hu2021lora}.
%

%
During the pre-training phase, we sample 500k images with segmentation labels from SA1B~\cite{kirillov2023segment}. For each image, we apply augmentations such as Crop, Resize, Flip, and Rotation to simulate a pseudo video. 
We then sample two frames from this pseudo video to serve as our training samples. Taking InternVL2~\cite{chen2024internvl} as the base model and RADIO~\cite{ranzinger2024radio} as the vision expert, we input one image into the InternVL2 visual encoder and the other into RADIO. When selecting the (anchor, positive, negatives) triplet, the anchor is chosen from the image features output by RADIO, while the positive and negatives are selected from the image features output by the InternVL2 visual encoder. We perform full training from scratch on the RADIO adapter using only object-level contrastive loss.

In the fine-tuning phase, we apply instruction augmentation to the original 220k MMVM data samples using object-level representations. Consequently, we utilize a total of 440k MMVM data samples during fine-tuning. When using Qwen2VL~\cite{wang2024qwen2} as the base model, to reduce sequence length and decrease computational resource requirements, we scale the long edge of all images to 1024 pixels and pad the short edge to 1024 pixels.

\noindent
\textbf{Inference details.}
When performing inference on the MMVM benchmark, we integrate CoLVA into the MLLMs. For inference on general VQA benchmarks, we maintain the MLLMs' original architecture and load the LLM parameters trained with CoLVA.

\section{More visualization results}
\label{sec:supp-more-qualitative-results}

\begin{figure*}[t]
\centering
\includegraphics[width=0.92\textwidth]{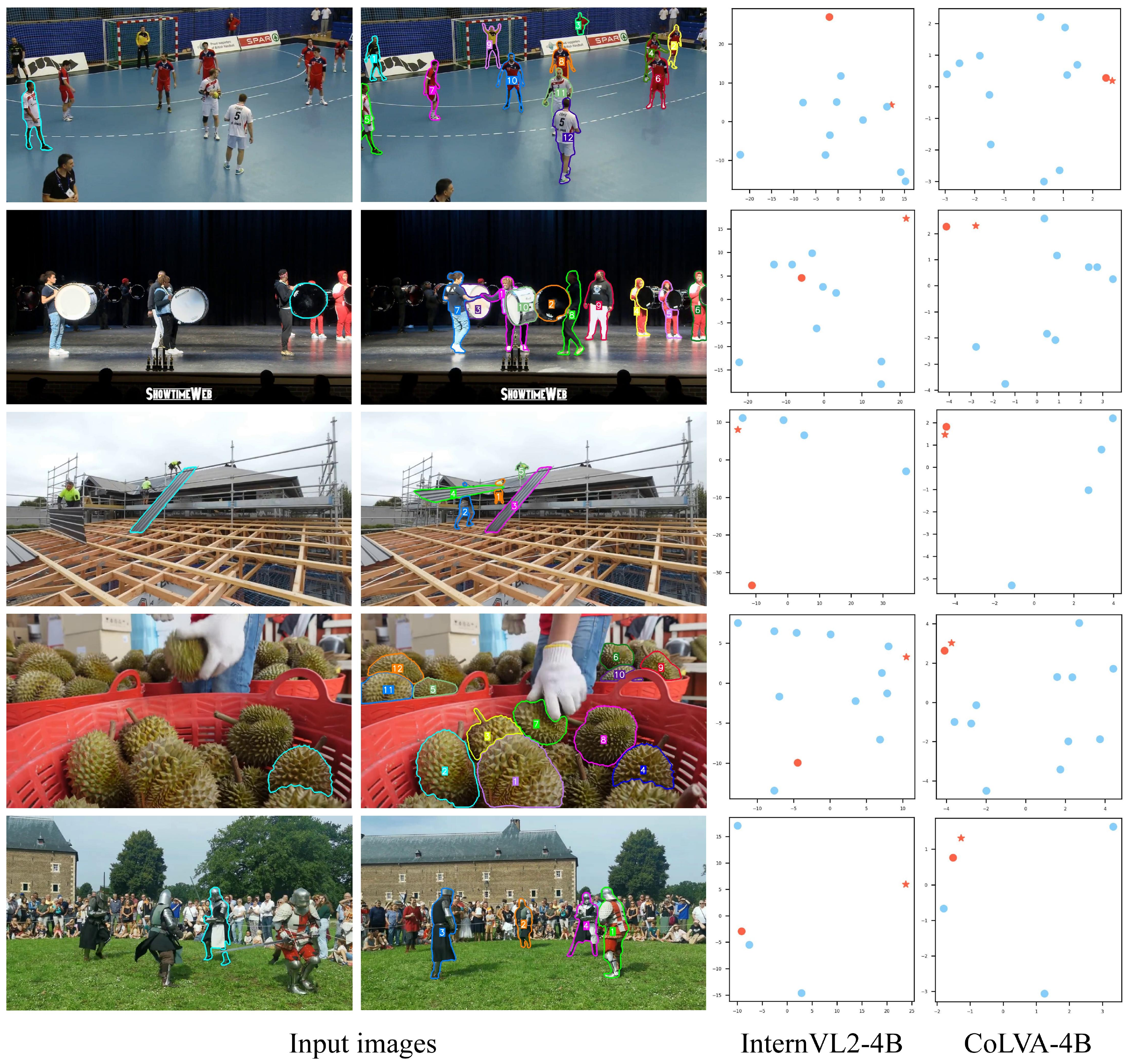}
\vspace{-2mm}\caption{\small More PCA visualizations of learned object embeddings by InternVL2-4B and our CoLVA-4B. The object embeddings are obtained by applying average pooling to the visual tokens using mask annotations. The red star represents the query object in the first image. The red dot represents the matched target in the second image. The blues dots represent other candidates. }
\label{fig:more_pca}
\end{figure*}

\noindent
\textbf{More PCA visualizations.}
In Fig.~\ref{fig:more_pca}, we present additional PCA visualizations. The results reveal that the matched target (represented by a red dot) and other candidate objects (represented by blue dots) are clustered together, while being distant from the query object (represented by a red star). This clustering pattern makes it challenging for InternVL2 to distinguish the correct object. In contrast, our CoLVA brings the matched target and the query object closer together while distancing them from other candidate objects. This indicates that our CoLVA has learned fine-grained and discriminative visual features, which are beneficial for visual matching tasks.

\begin{figure*}[t]
\centering
\includegraphics[width=0.92\textwidth]{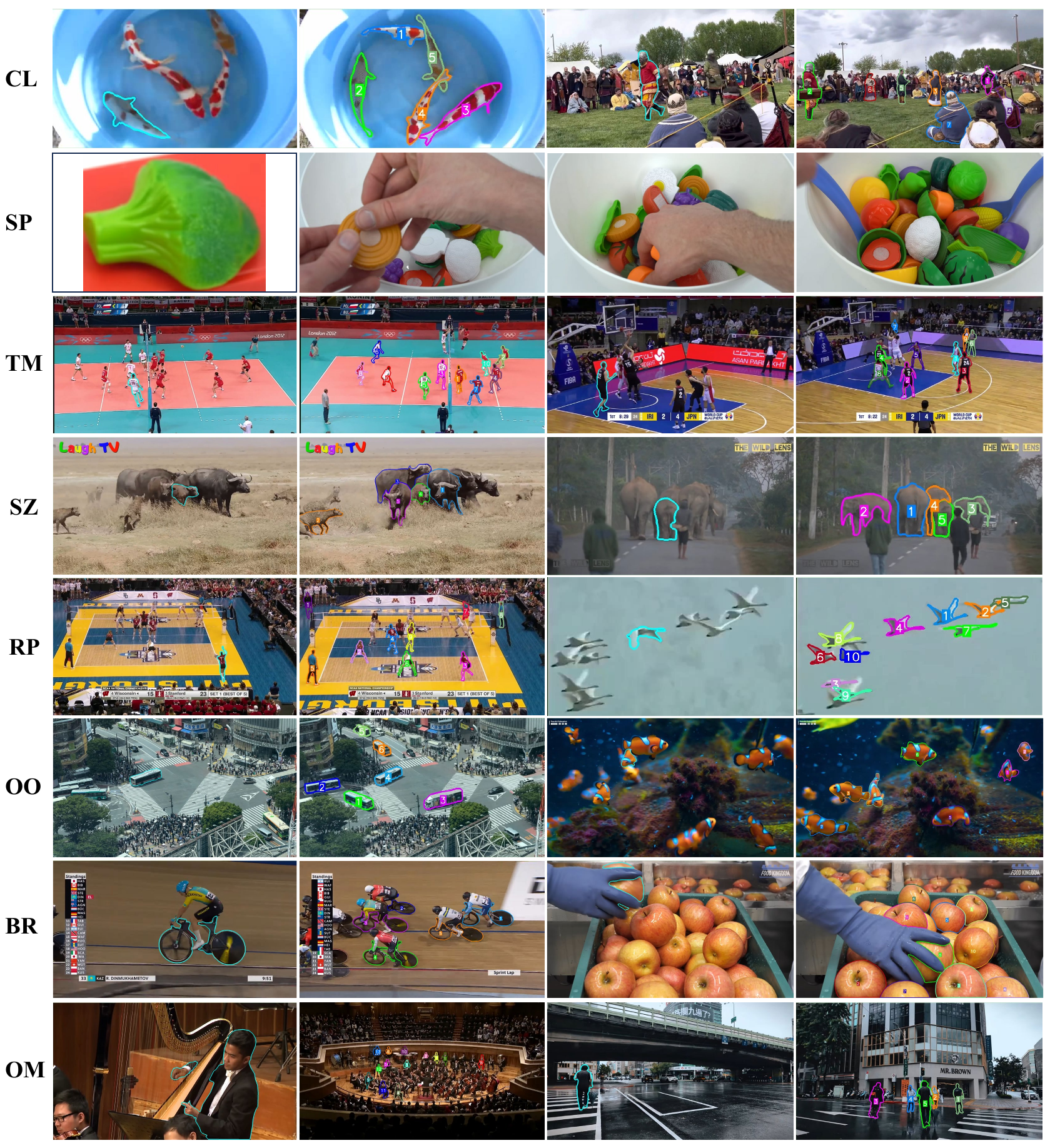}
\vspace{-2mm}\caption{\small More challenging test cases of our MMVM benchmark, where each row shows cases of different match types.}
\label{fig:mmvm_more_cases}
\end{figure*}

\noindent
\textbf{More challenging test cases of our MMVM.}
Here, we present more examples from the MMVM benchmark, which features diverse scenes and presents significant challenges, as illustrated in Fig.~\ref{fig:mmvm_more_cases}. In particular, our MMVM contains extremely small objects.

\section{Further Discussion}
\label{sec:supp-broader-impacts}

\noindent
\textbf{Future works.} We have argued the fine-grained visual perception and logical reasoning ability of MLLMs in the main paper. We give a more detailed description here. 

The former means the MLLMs must understand various scale objects well, where detailed information, such as object parts, remote objects, and thin objects, play a critical role in perception. Thus, equipping MLLMs with dense perception ability and visual prompts~\cite{Lai2023lisa,li2024omg,zhang2024omg,yuan2024osprey, ravi2024sam2} is needed.

The latter means that MLLMs must have instance-aware understanding and can perform visual comparisons~\cite{peng2024vasttrack}. With this ability, MLLMs can distinguish various objects and perform visual reasoning. This is why we adopt contrastive loss during the pre-training stage.

In addition, automatically collecting more high-quality supervised fine-tuning data is another way to boost MLLMs.

\begin{table*}[t!]
    \centering
    \caption{More MMVM Benchmark results. Accuracy is the metric, and the overall accuracy is computed across all 1,510 evaluation samples. The accuracy for each of the eight match types is calculated separately on their respective samples. The full term of the match type abbreviation can be found in the main text. For MLLMs that only support single-image input, we simply concatenate all the images vertically into one image and then input it.}\vspace{-2mm}
    \label{tab:more_match_bench_result}
    \resizebox{0.85\textwidth}{!}{
    \begin{tabular}{l | l | c c c c c c c c c}
    \toprule[1.5pt]
        Model Size & Method & Overall & CL & SP & TM & SZ & RP & OO & BR & OM \\
        \midrule[1pt]
        \multirow{8}{*}{$\sim$4B} & InternVL2-2B~\cite{chen2024internvl} & 9.87 & 9.66 & 6.90 & 10.28 & 10.39 & 8.28 & 11.20 & 10.80 & 8.80 \\
        ~ & xGen-MM-v1.5-4B~\cite{xue2024xgen} & 13.50 & 10.47 & 17.24 & 18.69 & 25.97 & 6.71 & 19.20 & 17.61 & 16.20 \\
        ~ & VILA1.5-3B\cite{lin2024vila} & 15.36 & 10.96 & 6.89 & 19.62 & 29.87 & 9.57 & 20.80 & 19.30 & 18.98 \\
        ~ & Qwen2-VL-2B-Instruct~\cite{wang2024qwen2} & 15.69 & 13.42 & 20.69 & 17.75 & 31.16 & 9.57 & 22.40 & 18.75 & 16.67 \\
        ~ & Ovis1.6-Llama3.2-3B~\cite{lu2024ovis} & 16.62 & 13.09 & 20.69 & 20.56 & 33.77 & 9.28 & 22.40 & 21.59 & 20.83 \\
        ~ & DeepSeek-VL-1.3B~\cite{lu2024deepseek} & 16.82 & 12.60 & 13.79	 & 18.69 & 37.66 & 10.43 & 22.40 & 21.59 & 17.59 \\
        ~ & InternVL2-4B~\cite{chen2024internvl} & 17.62 & 14.73 & 34.48 & 17.76 & 15.58 & 10.28 & 24.00 & 31.25 & 21.30 \\
        \midrule[1pt]
        \multirow{17}{*}{4B$\sim$13B} & Chameleon-7B~\cite{team2024chameleon} & 10.07 & 9.49 & 17.24 & 14.95 & 11.69 & 6.86 & 9.60 & 13.07 & 10.65\\
        ~ & Cambrian-13B~\cite{tong2024cambrian} & 10.72 & 9.32 & 6.89 & 9.34 & 23.37 & 6.28 & 16.00 & 15.34 & 7.87 \\
        ~ & Mini-Gemini-7B-HD~\cite{li2024mgm} & 13.18 & 10.80 & 10.34 & 14.95 & 25.97 & 8.28 & 14.40 & 18.18 & 13.89\\
        ~ & LLaVA-NEXT-13B~\cite{liu2024llavanext} & 13.77 & 8.35 & 10.34 & 10.28 & 22.08 & 7.57 & 22.4 & 22.73 & 18.52\\
        ~ & LLaVA1.5-13B~\cite{liu2023improvedllava} & 14.04 & 11.78 & 13.79 & 14.02 & 31.17 & 7.57 & 20.00 & 18.18 & 14.35\\
        ~ & MiniCPM-V2.5-8B~\cite{yao2024minicpmv} & 14.11 & 10.80 & 17.24 & 13.08 & 31.17 & 6.28 & 24.00 & 20.45 & 17.13\\
        ~ & Monkey-7B~\cite{li2023monkey} & 14.43 & 13.09 & 6.89 & 14.01 & 31.16 & 7.85 & 17.60 & 18.18 & 15.74 \\
        ~ & VILA1.5-13B~\cite{lin2024vila} & 14.70 & 13.91 & 13.79 & 13.08 & 36.36 & 7.57 & 22.40 & 17.04 & 15.74 \\
        ~ & Slime-13B~\cite{zhang2024beyond} & 14.83 & 11.29 & 6.89 & 16.82 & 32.46 & 9.00 & 18.40 & 21.02 & 17.59 \\
        ~ & mPLUG-Owl3-7B~\cite{ye2024mplug} & 16.22 & 14.07 & 20.68 & 16.82 & 31.16 & 8.57 & 20.80 & 20.45 & 19.90 \\
        ~ & InternVL2-8B~\cite{chen2024internvl} & 16.89 & 13.58 & 20.69 & 22.43 & 24.68 & 11.57 & 24.00 & 23.30 & 18.52 \\
        ~ & VITA-8*7B~\cite{fu2024vita} & 17.42 & 14.57 & 13.79 & 23.36 & 29.87 & 10.57 & 24.80 & 22.16 & 20.37\\
        ~ & DeepSeek-VL-7b~\cite{lu2024deepseek} & 17.68 & 14.24 & 17.24 & 20.56 & 35.06 & 10.00 & 22.40 & 25.00 & 23.61\\
        ~ & Ovis1.6-Gemma2-9B~\cite{lu2024ovis} & 17.75 & 17.68 & 17.24 & 15.89 & 32.47 & 12.14 & 20.00 & 19.32 & 18.98\\
        ~ & LLaVA-Next-Interleave-7B~\cite{li2024llavanextinter} & 19.34 & 15.88 & 41.38 & 15.89 & 41.56 & 10.71 & 19.20 & 23.30 & 27.78 \\
        ~ & LLaVA-OneVision-ov-7B~\cite{li2024llavaonevision} & 20.92 & 16.69 & 17.24 & 25.23 & 31.16 & 14.28 & 22.40 & 30.68 & 25.92\\
        ~ & Qwen2-VL-7B-Instruct~\cite{wang2024qwen2} & 27.48 & 24.87 & 37.93 & 30.84 & 62.33 & 17.85 & 28.00 & 28.97 & 31.94\\
        \midrule[1pt]
        \multirow{6}{*}{13B$\sim$40B} & Yi-VL-34B~\cite{young2024yi} & 11.26 & 9.49 & 17.24 & 18.69 & 12.99 & 7.57 & 9.60 & 15.34 & 11.57 \\
        ~ & Eagle-X5-34B-Chat~\cite{shi2024eagle} & 13.84 & 10.47 & 13.79 & 13.08 & 27.27 & 7.86 & 23.20 & 18.18 & 14.81 \\
        ~ & LLaVA-Next-34B~\cite{liu2024llavanext} & 15.03 & 11.29 & 20.69 & 16.82 & 32.47 & 8.71 & 21.6 & 19.89 & 17.13 \\
        ~ & VILA1.5-40B~\cite{lin2024vila} & 15.36 & 14.73 & 20.69 & 14.95 & 36.36 & 5.00 & 22.40 & 18.18 & 17.13 \\
        ~ & InternVL2-40B~\cite{chen2024internvl} & 26.03 & 24.88 & 41.38 & 33.64 & 42.86 & 16.86 & 31.20 & 31.82 & 31.02\\
        \midrule[1pt]
        \multirow{5}{*}{40B$\sim$} & Idefics-80B-instruct~\cite{laurenccon2024obelics} & 13.58 & 11.13 & 13.79 & 14.95 & 24.68 & 7.00 & 20.80 & 	17.61 & 13.89 \\
        ~ & InternVL2-76B~\cite{chen2024internvl} & 25.83 & 24.06 & 31.03 & 30.84 & 40.26 & 19.28 & 31.20 & 30.11 & 31.02 \\
        ~ & LLaVA-OneVision-ov-72B~\cite{li2024llavaonevision} & 29.34 & 28.48 & 34.48 & 26.17 & 55.84 & 21.14 & 28.00 & 34.66 & 32.41 \\
        ~ & InternVL2.5-78B~\cite{chen2024internvl2_5} & 36.42 & 35.02 & 37.93 & 38.32 & 58.44 & 25.86 & 38.40 & 39.20 & 43.98 \\
        ~ & Qwen2-VL-72B-Instruct~\cite{wang2024qwen2} & 38.08 & 37.64 & 44.83 & 42.06 & 64.94 & 32.28 & 36.00 & 35.80 & 39.81\\
        \midrule[1pt]
        \multirow{3}{*}{Unkown} & Claude3-5V-Sonnet & 40.20 & 34.21 & 41.38 & 56.07 & 77.92 & 34.86 & 40.00 & 32.39 & 40.28 \\
        ~ & GeminiPro1-5 & 40.73 & 36.00 & 44.83 & 44.86 & 74.02 & 35.14 & 44.80 & 38.07 & 38.42 \\
        ~ & GPT4o-20240806 & 42.65 & 39.28 & \textbf{65.52} & \textbf{60.75} & 67.53 & 32.28 & 44.00 & 43.18 & 50.00 \\
        \midrule[1pt]
                2B & CoLVA-Qwen2VL-2B (Ours) & 47.48 & 40.92 & 31.03 & 47.66 & 68.83 & \textbf{50.57} & 49.60 & 33.52 & 38.42 \\
                4B & CoLVA-InternVL2-4B (Ours) & 49.80 & \textbf{43.21} & 41.38 & 45.79 & 77.92 & 44.43 & \textbf{53.60} & 44.89 & \textbf{53.24} \\
                7B & CoLVA-Qwen2VL-7B (Ours) & \textbf{51.06} & 42.72 & 37.93 & 49.53 & \textbf{80.52} & 46.43 & 52.80 & \textbf{47.73} & 49.54 \\
    \bottomrule[1.5pt]
    \end{tabular}
    }\vspace{-4mm}
\end{table*}

\noindent
\textbf{Board impact.} Our works explore one fundamental limitation of current SOTA MLLMs: visual correspondence shortcomings. We present a new benchmark: MMVM, a training dataset, and a new training framework, CoLVA, to improve the visual correspondence in MLLM models. Our work will raise the attention of visual correspondence in MLLM design and inspire research on cross-image VQA tasks and fine-grained VQA tasks.
